# Enhanced hermit crabs detection using super-resolution reconstruction and improved YOLOv8 on UAV-captured imagery

Fan Zhao *, Yijia Chen, Dianhan Xi, Yongying Liu, Jiaqi Wang, Shigeru Tabeta, Katsunori Mizuno *

*Graduate School of Frontier Sciences, The University of Tokyo, Japan*

ABSTRACT

Hermit crabs are vital to coastal ecosystems, serving as environmental health indicators and contributing to seed dispersal, debris cleanup, and soil disturbance. Traditional hermit crabs survey methods, like quadrat sampling, are labor-intensive and environmentally dependent. This study presents an innovative approach combining UAV (Unmanned Aerial Vehicles)-based remote sensing with Super-Resolution Reconstruction (SRR) and the CRAB-YOLO detection network, a modification of YOLOv8s, to monitor hermit crabs effectively. SRR enhances image quality by addressing motion blur and insufficient resolution, significantly improving detection accuracy over conventional low-resolution fuzzy images. CRAB-YOLO integrates three improvements for accuracy, hermit crab characteristics, and computational efficiency, achieving state-of-the-art (SOTA) performance. The Residual Dense Network (RDN) demonstrated the best image reconstruction performance, and CRAB-YOLO achieved a mean average precision (mAP) of 69.5 % on the SRR test set, a 40 % improvement over the conventional Bicubic method with a magnification factor of 4. These results prove the effectiveness of the proposed method for cost-effective, automated hermit crab monitoring supporting coastal benthos conservation efforts.

## 1. Introduction

### 1.1. Importance of hermit crabs in marine ecosystems

Hermit crabs belong to the superfamily *Paguroidea* in the order *Decapoda*, *subphylum Crustacea*, and *phylum Arthropoda*. They significantly impact coastal ecosystems by aiding seed dispersion and cleaning up debris, accelerating organic matter decomposition (Sant'Anna et al., 2012). Hermit crabs also disturb soil through burrowing and gather fallen leaves (Lavers et al., 2020). Furthermore, they play a crucial role in nature conservation. In response to climate change, hermit crabs exhibit behaviors related to global warming (Asakura, 2021), ocean acidification (de la Haye et al., 2012; Turra et al., 2020), eutrophication (Briffa et al., 2023), organic pollutants (Sant'Anna et al., 2012), and plastics (Briffa et al., 2023; Lavers et al., 2020). Additionally, hermit crabs can act as indicator species due to their sensitivity to pollutants such as heavy metals (Nafchi and Chamani, 2019). Conducting large-scale surveys of hermit crabs is essential for understanding their population dynamics and health, providing valuable insights into marine ecosystem conditions.

### 1.2. Challenges and advancements in remote sensing for hermit crab monitoring

Traditional survey techniques for benthos like hermit crabs include quadrat sampling (Briffa et al., 2023), are time-consuming and labor-intensive. Environmental variables like water clarity, depth, and underwater terrain can also affect their reliability. Large-scale population-level surveys of hermit crabs using traditional methods usually require substantial resources. Additionally, these methods often retail manually setting up sampling areas, implying that conducting surveys beyond the intertidal zone typically demands expensive vessels. This approach is often unaffordable for remote islands and other underfunded organizations and regions.

Recent advancements in remote sensing technology have introduced acoustic, satellite, and airborne remote sensing techniques for surveying benthos. Acoustic techniques including side-scan sonar (Matarrese et al., 2004; Shih et al., 2019), multibeam sonar (Brehmer et al., 2006), and echo sounders (Brown et al., 2019), help overcome limitations posed by water clarity. Adaptive resolution imaging sonar (ARIS) is notable for visualizing turbid water environments (Zhao et al., 2022, 2023).

* Corresponding authors.
*E-mail addresses:* zhaofan25ut@163.com, zhaofan@edu.k.u-tokyo.ac.jp (F. Zhao), kmizuno@edu.k.u-tokyo.ac.jp (K. Mizuno).

However, its low image resolution hampers the ability to discern fine details or accurately identify organisms within the surveyed area. Moreover, the high cost of acoustic surveys presents a challenge for researchers or organizations with limited resources.

UAV-based remote sensing is a powerful and cost-effective solution for monitoring marine ecosystems (Kieu et al., 2023). UAVs are useful for monitoring benthos such as scallops, sea cucumbers, and clams (Geraeds et al., 2019). When combined with computer vision technologies, UAVs can effectively identify benthos. Compared to acoustic and satellite remote sensing methods, UAV-based technology is more cost-effective and efficient (Ventura et al., 2018). Consumer-grade drones, in particular, offer substantial potential in marine biology research, aiding in the protection of marine ecosystems.

### 1.3. SRR in enhancing UAV images

The efficiency of UAV-based remote sensing depends on the quality of the images obtained (Colefax et al., 2018). However, UAVs are prone to vibration, leading to motion blur and suboptimal resolution (Liu et al., 2020). Additionally, environmental factors such as water depth, light reflection on the water surface, and water turbidity significantly affect image quality. Higher altitudes during drone flights increase survey efficiency but reduce image resolution. These factors limit the quality of image data collected by UAVs, making them unsuitable for detailed ecological analysis. Therefore, implementing efficient image processing technology is essential to address these challenges in UAV-based ecological monitoring. Conventional image processing techniques, such as unsharp mask filtering, median filtering, and histogram equalization (Bae et al., 2021), enhance image quality based on existing pixel information rather than improving image resolution, offering limited value for detailed analysis (Zhao et al., 2023). SRR is expected to address UAV issues like motion blur and inadequate image resolution (Benecki et al., 2018). It includes three algorithmic categories: interpolation, reconstruction, and machine learning (Ooi and Ibrahim, 2021). Each SRR method has drawbacks. Interpolation-based methods, which focus on pixel manipulation, often result in blurred images due to excessive detail reduction. Reconstruction-based approaches integrate prior knowledge but fall short in reconstructing texture-rich images (Wang et al., 2015). Traditional machine learning algorithms provide more precise outcomes but are time-intensive and challenging to optimize (Zhao and Zhang, 2018).

In contrast, deep learning-based SRR algorithms yield more accurate results and excel in advanced image processing tasks across various fields (Yue et al., 2016). These algorithms have found successful applications in diverse domains, including medical imaging (Yang et al., 2023; Mahapatra et al., 2019), object detection (Jin et al., 2021), and face recognition (Rasti et al., 2016). SRR enhances the quality of low-resolution images acquired by UAVs by improving image resolution, thus proving its potential in creating more precise and reliable models from UAV-captured images (Wang et al., 2024; González et al., 2022; Zhao et al., 2024a, 2024b). Jin et al. (2021) and Zhao et al. (2025a, 2025b) found that SRR not only enhances image resolution but also improves the segmentation and detection accuracy of specific objects. Xiang et al. (2022) applied SRR to process images of cracks on building peripheries taken by UAVs, improving visual and quantitative image quality and increasing the IoU of the improved semantic segmentation model by about 16 %. Therefore, SRR can refine the quality of optical images, including UAV-captured images, providing more details for ecological evaluation and monitoring.

### 1.4. Computer vision in marine ecology: object detection and quantification

In the field of computer vision, convolutional neural networks (CNNs) have significantly advanced automated image classification and object detection tasks (Dhillon and Verma, 2020). This progress is crucial for ocean ecology by enabling the quantification of marine life and objects. For instance, Fast-CNN can achieve an accuracy of 96.32 % on a small marine benthos image dataset, demonstrating the effectiveness of CNN in quantifying marine benthic organisms (Liu and Wang, 2021). CNN-based object detection models facilitate the automated detection and quantification of marine benthic organisms, thereby providing efficient and cost-effective assessment and monitoring of the marine ecological environment.

Despite their high accuracy, these models often lack the capability to locate objects within an image, which is essential for benthos object recognition and subsequent population density studies. The You Only Look Once (YOLO) series networks have been widely used in object detection tasks across various fields, including autonomous driving, medical imaging, civil engineering, and security monitoring. YOLO networks include location information, making them promising for addressing this issue. Xu et al. (2023) improved the YOLOv5 model for detecting dense small-scale marine benthic organisms, achieving excellent results. Fu et al. (2022) proposed an improved YOLOT for underwater benthos object detection, enabling the quantification of underwater benthos based on detection results. This innovation enhanced detection speed, allowing for swift object identification and position determination. The crucial information provided by the YOLO series models has significantly contributed to research on underwater benthos ecological monitoring, making them well-known models in the field. In this study, YOLOv8, the classic iteration of the YOLO series, is used for the detection of hermit crabs, which were integrated with UAVs to monitor hermit crabs.

### 1.5. Research Objective and contributions

This research aims to develop a method for large-scale detection of hermit crabs in UAV images by combining SRR networks for image enhancement with YOLOv8 for object detection (Zhao et al., 2024c). The main contributions of this work are as follows: (1) a deep learning-based SRR framework to enhance UAV image quality, (2) the development of CRAB-YOLO, a modified YOLOv8 network optimized for hermit crab detection, (3) a comprehensive evaluation of SRR algorithms and object detection models, and (4) a quantitative assessment of detection accuracy under varying magnification factors. The process includes training SRR networks using diverse algorithms and evaluating the quality of reconstructed hermit crab images through theoretical metrics and visual comparisons. The proposed object detection network, CRAB-YOLO, is a modified version of YOLOv8 specifically optimized for hermit crab detection. It incorporates three key improvements: enhanced detection accuracy, adaptation to hermit crab characteristics, and increased computational efficiency. CRAB-YOLO is trained on high-resolution hermit crab datasets to identify hermit crabs in test sets. Detection results are compared and analyzed in detail, including a quantitative assessment of the object detection model's performance. The detection results of the proposed CRAB-YOLO were compared with those of state-of-the-art models, including Fast-CNN, SSD, and other YOLO series models, to demonstrate its superior performance in hermit crab detection. Furthermore, the detection accuracy of hermit crab images reconstructed at different magnification factors is thoroughly examined.

The structure of this article is organized as follows: Section 2 illustrates the flow of the proposed method, in the order of the process of SRR, the architecture of CRAB-YOLO, and the quantitative assessment of SRR algorithms and object detection networks. Section 3 proposed the implementation details and dataset preparation, followed by the presentation of experimental results using various SRR networks and object detection networks, along with a comprehensive comparative evaluation. Section 4 discusses the effects of key parameters on the results, culminating in the conclusions outlined in Section 5.

## 2. Methodology

Fig. 1 presented the workflow of the proposed approach, consisting of three primary phases: (1) UAV-captured images of hermit crabs, which include blurred or low-resolution (LR) images, are reconstructed into SR images using various deep learning-based SRR models. These models are trained to study the mapping function between LR and high-resolution (HR) images through a training dataset, enabling them to generate enhanced super-resolved images (SR images) from new LR inputs. (2) In the second phase, object detection is performed on the SR images of hermit crabs reconstructed from the first phase based on proposed YOLOv8 model, i.e. CRAB-YOLO. The outputs from different SRR models serve as inputs to CRAB-YOLO and are compared both theoretically and visually. (3) The impact of SRR on hermit crab detection accuracy is quantitatively evaluated based on the results from the second phase. Detailed findings are presented in Sections 2.1 to 2.3.

### 2.1. Survey site

Lake Hamana, a brackish lagoon in Shizuoka Prefecture, Japan, lies near the Pacific coast (34.7411°N, 137.5697°E) (Fig. 2). Connected to the Pacific Ocean via a narrow channel, the lake experiences water level variations of up to 1.2 m, with over 40 million tons of seawater shifting between high and low tides. Covering 65 km$^2$ with a coastline of 114 km, Lake Hamana hosts diverse aquatic benthos, including hermit crabs, oysters, and soft-shelled turtles. It serves as a crucial resource for monitoring the impacts of climate change, eutrophication, and human activities on its ecosystem, functioning also as a living laboratory for studying the ecology and environment of brackish lakes.

### 2.2. Data collection

The DJI Phantom 4 drone was selected for its stability, high-resolution imaging capabilities, and extended flight times of up to 28 min for this study. The camera was positioned at nadir with a field of view (FOV) of 94° and a pixel array of 5472 x 3648. The UAV operated at a height of 5 m. Data collection was timed to coincide with low tide periods, typically around 3PM, to minimize sun glint and ensure optimal visibility. The survey encompassed depths ranging from exposed lake-beds to water depths between 10 cm and 30 cm. Special care was taken to avoid areas of human interference or open habitats that could influence the natural behavior of hermit crabs. The entire data collection process spanned approximately half an hour.

### 2.3. Data preprocessing

The data preprocessing phase begins with cropping drone-captured imagery using the sliding window technique. A window of size 640 × 640 pixels moves across the entire image with a stride of 320 pixels. At each step, the section of the image within the window is extracted, resulting in a series of smaller image patches. This approach generates 1720 cropped images from each original high-resolution drone image. Subsequently, these HR images undergo downsampling to 160x160 pixels using the degradation model described in Section 2.1 to create LR images. The dataset is divided into training, validation, and test subsets following an 8:1:1 ratio distribution. Following cropping and down-sampling, image augmentation is performed. This includes rotating images by 180° and applying horizontal and vertical flipping transformations, as well as scaling images by ratios of 0.6, 0.7, 0.8, and 0.9. These augmentations not only diversify visual perspectives within the dataset but also expand the dataset size by a factor of 30.

### 2.4. Super-resolution reconstruction

Deep learning-based SRR is often used to enhance the resolution of low-quality images. The core principle involves training a model to learn the mapping between LR and HR images, allowing it to generate high-quality outputs (SR images) from LR inputs. The process typically begins with dataset preparation, where paired HR images are collected and LR counterparts are derived from them. The selection of an SRR model is critical, as different models based on various architectures and design strategies offer varying capabilities in handling image features such as edges and textures, thus impacting the overall performance of object detection models. During training, the LR images are fed into the SRR model, which is optimized using loss functions that quantify the discrepancy between the predicted SR images and the actual HR images. Key parameters influencing SRR performance include the choice of SRR algorithm, the size and quality of the training dataset, and the neural network architecture. These factors are pivotal in determining the model's ability to accurately reconstruct fine details and enhance image quality. The experiment environment parameters are shown in Table 1. For model training, the input image size was set to 640 × 640 pixels, with an initial learning rate of 0.01. The learning rate was updated using stochastic gradient descent with a momentum of 0.937 and weight decay of 0.0005. Data augmentation techniques, including Mosaic augmentation, were applied to enrich the training dataset and improve model generalization. Each training batch consisted of 8 images, processed by 8 worker threads. The models were trained for 200 epochs, with label smoothing set to 0.01 to prevent overfitting. Following training, the effectiveness of the SRR model is evaluated using metrics, alongside a qualitative visual comparison.

#### 2.4.1. Architecture of SRR networks

Numerous SRR models have been proposed by researchers using various network architectures. In this research, five typical networks aligned with distinct network strategies were chosen to formulate SRR models based on the experience of similar studies and the properties of the hermit crab object detection task. The architectures of these networks are depicted in Fig. 3, where "Conv" indicates the convolutional layer and "Deconv" represents the deconvolutional layer.

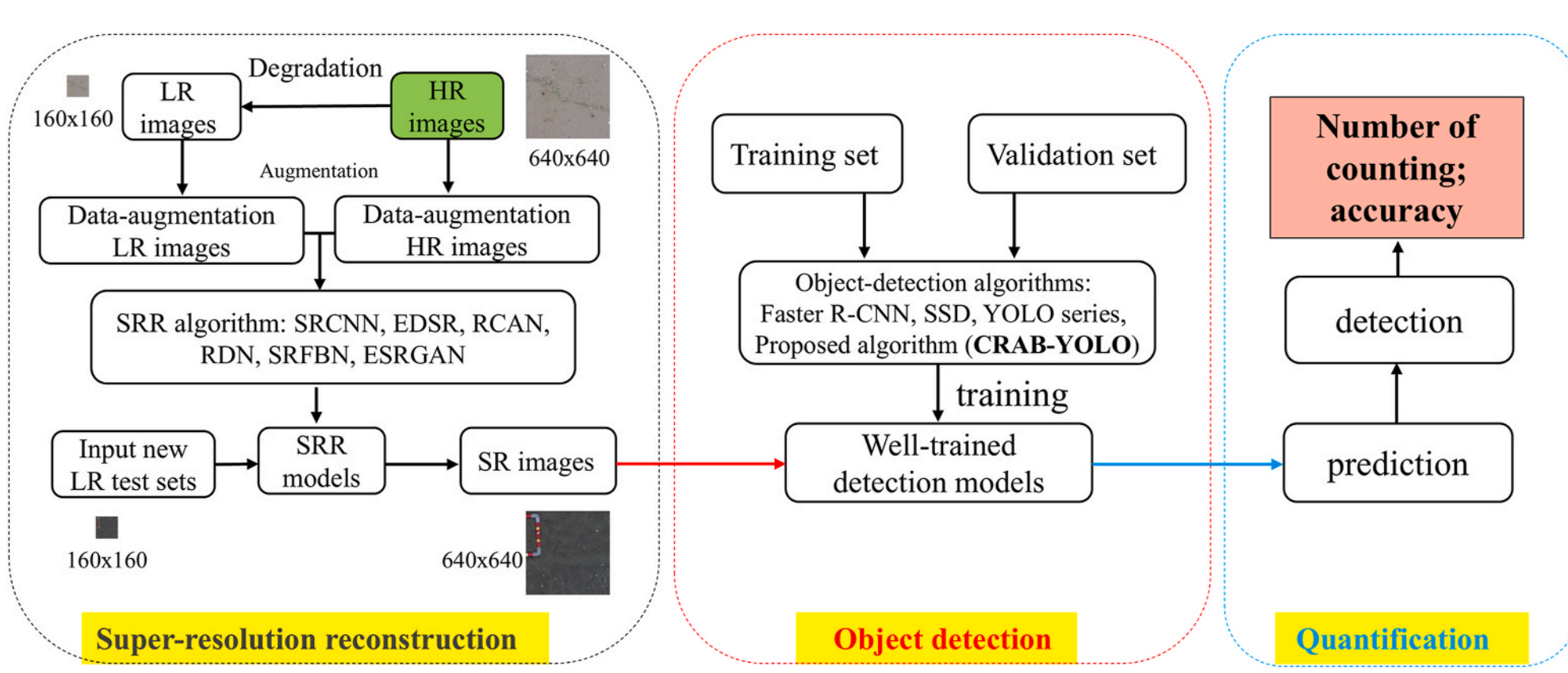

**Fig. 1.** Workflow of the proposed approach for the monitoring of hermit crabs using UAV images.

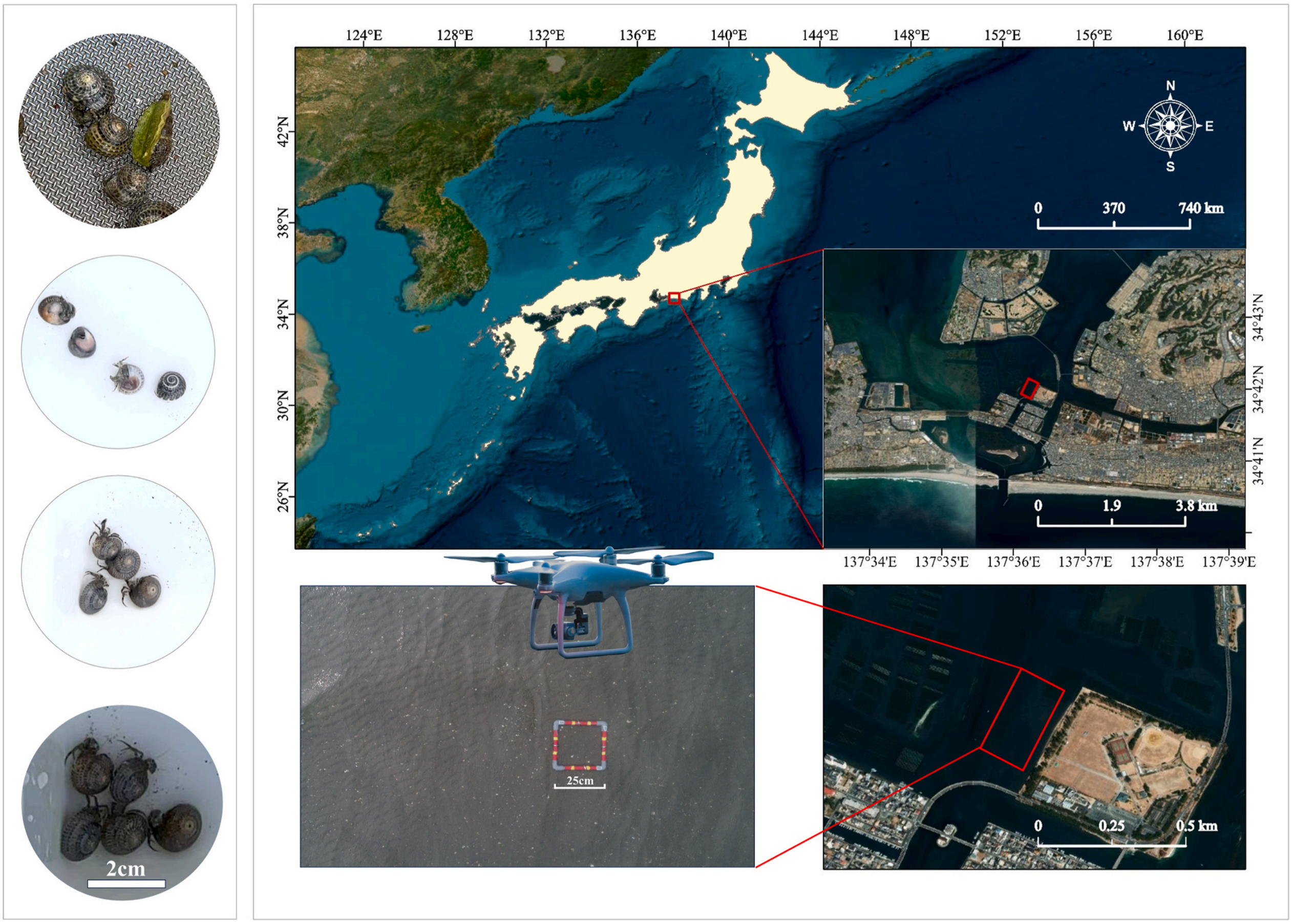

**Fig. 2.** Map of survey site at Lake Hamana, Shizuoka Prefecture, Japan, illustrating the drone flight process.

**Table 1**
Experimental environment parameters.

| Experiment configuration | Type or version |
|---|---|
| Operating system | Ubuntu 18.04 |
| Bit architecture | 32 bit |
| Random Access Memory | 64 GB |
| GPU | NVIDIA GeForce RTX 4090 |
| Memory | 24 GB |
| CPU | Intel(R) Xeon(R) Platinum i9-13900k |
| Pytorch | 1.10 |
| CUDA | 11.1 |
| Cudnn | 8.0.4 |

Super-Resolution Convolutional Neural Network (SRCNN) is a widely used deep learning model for SRR known for its simplicity. It features a straightforward architecture with just three layers: image feature extraction, non-linear mapping, and image reconstruction. This uncomplicated design makes SRCNN easy to implement and train, though it may not capture complex image features as effectively as more advanced models. EDSR stands out by excluding the Batch Normalization (BN) layer, which reduces memory consumption by approximately 40 % during training. This reduction allows for more network layers to be stacked, enabling the extraction of richer features. Consequently, Enhanced Deep Super-Resolution (EDSR) is highly efficient and capable of producing superior high-resolution images. Residual Channel Attention Networks (RCAN) increases network depth while dynamically focusing on relevant channel features (Zhang et al., 2018a). It excels in bypassing low-frequency information and emphasizing high-frequency details. The channel attention mechanism allows RCAN to adjust features based on their interdependencies, achieving accurate restoration, which can be proved in various empirical results. RDN combines the strengths of residual learning and dense connections. In this model, features from each layer are combined and reused, enhancing the information available for image reconstruction (Zhang et al., 2018b). This dense connectivity effectively addresses the challenge of gradient vanishing and maintains model performance without significantly increasing its size. The Super-Resolution Feedback Network (SRFBN) introduces a feedback mechanism to enhance deep learning-based image super-resolution. SRFBN refines low-level representations iteratively with high-level information (Li et al., 2019). A feedback block manages these connections, producing powerful high-level representations and superior early-stage reconstructions. The use of a curriculum learning strategy further improves SRFBN's adaptability to complex tasks, demonstrating its superiority in achieving high performance with fewer parameters.

### *2.4.2. Training of the networks*

During the training process, model parameters are optimized iteratively by adjusting them based on the computed loss between the input LR image and its corresponding HR counterpart. The optimization is achieved using the L1 loss function, formally defined as:

$$\mathscr{L}_1(O,R) = \frac{1}{pq}\sum_{i=0}^{p-1}\sum_{j=0}^{q-1} \| O(i,j) - R(i,j) \| \tag{1}$$

Here, $R(i,j)$ denotes the SR image, whereas $O(i,j)$ signifies the original HR image with a resolution of $p \times q$. SRR models were chosen based on their architectural diversity, effectiveness in handling various aspects of image super-resolution, and suitability for the specific requirements of hermit crab detection tasks. SRCNN serves as a straightforward baseline model, while EDSR's enhanced feature extraction capabilities are crucial for detailed image reconstruction. RDN's dense connections and RCAN's focus on high-frequency details address the challenges in complex feature mapping and fine detail preservation, and SRFBN's feedback mechanism and iterative refinement are ideal for incremental improvements in image quality.

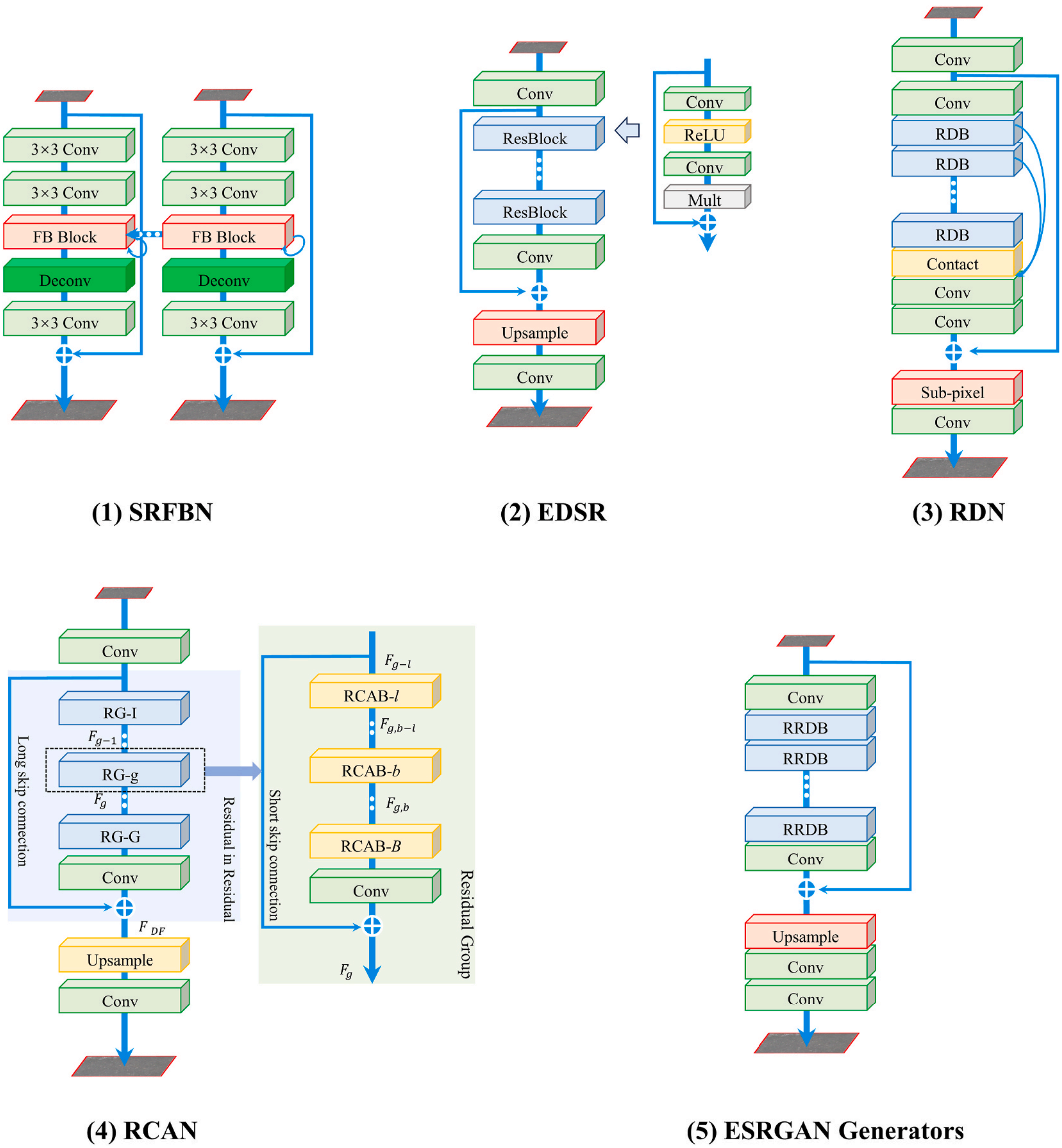

**Fig. 3.** Network Architectures of selected super-resolution reconstruction.

### 2.4.3. Evaluation metrics

The effectiveness of reconstruction was assessed using metrics, namely Peak Signal-to-Noise Ratio (PSNR), and Structural Similarity Index (SSIM) (Ooi and Ibrahim, 2021; Ward et al., 2017). PSNR, as defined in Equation (2) as:

$$PSNR = 10 \times \lg\left(\frac{{MAX_I}^2}{MSE}\right) \tag{2}$$

Here, PSNR quantifies the similarity between SR and original HR images. $MAX_I$ represents the highest gray value within the image, typically set to 255. A higher PSNR value indicates greater similarity. On the other hand, SSIM serves as an evaluative criterion for image quality, considering factors like brightness, contrast, and structural attributes to assess the similarity between two images. A SSIM score of 1 indicates perfect congruence between the generated and original images. The mathematical expressions for SSIM, outlined in Equation (3):

$$SSIM(O,R) = \frac{(2\mu_O\mu_R + C_1)(2\sigma_{OR} + C_2)}{(\mu_O^2 + \mu_R^2 + C_1)(\sigma_O^2 + \sigma_R^2 + C_2)} \tag{3}$$

Here, $\mu_O$ and $\mu_R$ represent the means of the total pixels in images $O(i,j)$ and $R(i,j)$, respectively. Similarly, $\sigma_O$ and $\sigma_R$ denote the variances of images $O$ and $R$, while $\sigma_{OR}$ denotes their covariance. Constants $C_1$ and $C_2$ are introduced to ensure numerical stability in the calculation, preventing division by zero in the denominator.

## 2.5. Object detection

Underwater object detection in computer vision poses significant challenges due to the complex underwater environment, resulting in degraded images with high noise, low visibility, and color deviation. Traditional object detection methods struggle with accuracy and generalization in this domain. The creators of YOLO (Redmon et al., 2016) revolutionized object detection by framing it as a regression problem, departing from traditional classification approaches. YOLO predicts class probabilities and delineates bounding boxes encompassing all objects in an image in a single step. This operational efficiency led to the acronym "You Only Look Once" (YOLO), highlighting its capability for rapid and comprehensive object recognition. YOLO networks are renowned for their real-time processing capability, making them suitable for time-sensitive applications. In addition to speed, YOLO exhibits high detection accuracy and generalization across diverse scenarios, attributed to its use of a single convolutional neural network for comprehensive image processing (Jiang et al., 2022).

### 2.5.1. YOLOv8 object detection network

YOLOv8 stands out in object detection due to its single-stage architecture, ensuring real-time processing by detecting objects in a single pass, this approach significantly enhances speed and efficiency compared to previous versions. Its anchor-free detection and improved feature pyramids contribute to enhanced accuracy, particularly for small and obscure underwater objects such as hermit crabs. The model's robustness across various scales and aspect ratios, achieved through refined training and data augmentation techniques, is crucial for the diverse underwater environment. YOLOv8 utilizes an efficient backbone and advanced feature pyramid network that enhance feature extraction and fusion, resulting in superior performance even in challenging conditions such as low visibility and varying lighting (Shen et al., 2023). These attributes establish YOLOv8 as the preferred backbone model for this study.

### 2.5.2. Architecture of proposed CRAB-YOLO

While YOLOv8 offers numerous practical advantages, it encounters limitations in detecting hermit crabs in UAV images. These limitations arise from the model's inherent design, which may not adequately preserve fine details of smaller objects during the feature extraction process in deep neural networks, thereby compromising crucial detection accuracy. Moreover, UAV imagery often includes complex backgrounds and numerous small-scale objects, exacerbating these challenges (Chia et al., 2023). To address these shortcomings of YOLOv8, this study enhances its object detection capabilities in three ways, leading to the development of the CRAB-YOLO model, as shown in Fig. 4.

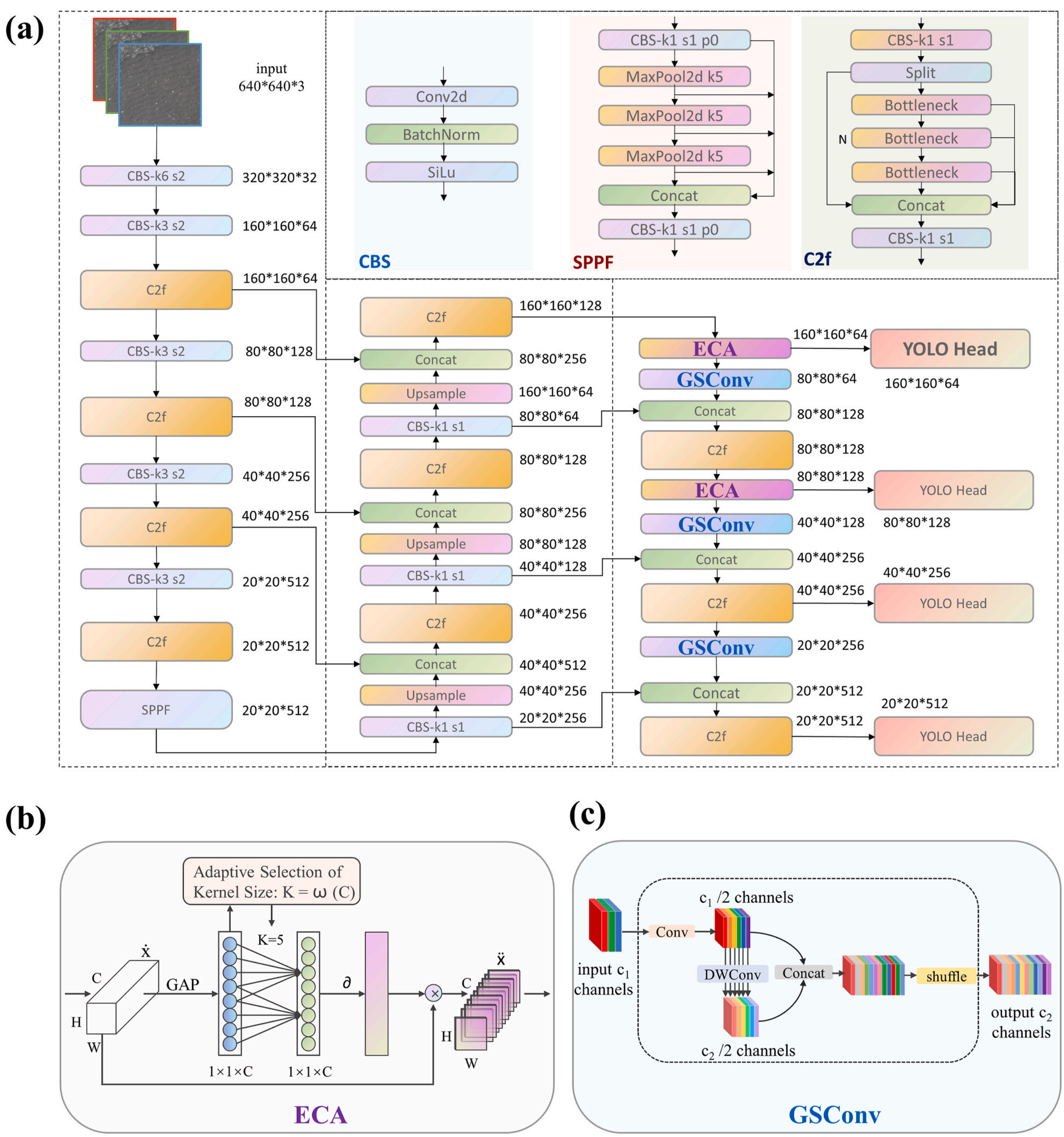

**Fig. 4.** (a) Enhanced modules and network structure of the CRAB-YOLO model; (b) architecture of the ECA attention mechanism; (c) GSConv module configuration.

The first enhancement is the composition of four detection heads to improve feature extraction efficiency. YOLOv8 typically uses three detection heads, but this study improves it by employing four detection heads, which aims to fully exploit the information of objects within lower-level feature maps, thereby improving the detection of small objects in complex backgrounds and enhancing overall feature extraction efficiency. The CRAB-YOLO also introduces the GSConv network structure, which enhances feature extraction capabilities. GSConv modules are integrated into layers 11 and 12, improving model accuracy without increasing the overall parameter count. This enhancement boosts overall feature extraction while maintaining balanced computational load. Additionally, the ECA mechanism integrated within the network enhances feature representation and fusion, making the model more effective for detecting hermit crabs in UAV imagery. Additional detection head in CRAB-YOLO object detection tasks enhances the model's ability to extract and exploit detailed features. This improvement is meaningful because it allows the model to capture finer details from lower-level feature maps in UAV images, which are essential for accurately identifying hermit crabs that might otherwise be overlooked. Although adding detection heads may increase additional computational effort, results from similar studies show that this expenditure is necessary (Song et al., 2020). UAV imagery often presents complex backgrounds and numerous small objects, making accurate detection challenging. Hermit crabs, being small and often camouflaged against their environment, require a model that can distinguish them from the background. By using four detection heads, the CRAB-YOLO can better focus on these small features, improving the precision and reliability of detecting hermit crabs, which not only boosts detection accuracy but also ensures that critical details are preserved, leading to more robust and effective monitoring of hermit crab.

The GSconv convolution module is an innovative lightweight convolution technique designed to enhance CNN efficiency. The GSconv module in this research is composed of a depth-wise separable convolution layer and a transposed convolution layer (Li et al., 2022). In this research, GSConv structures are added at layers 11 and 12, boosting model accuracy without increasing the overall parameter count. This module excels in efficient feature extraction, crucial for maintaining high performance in computationally limited scenarios like real-time UAV processing. For hermit crab detection in UAV imagery, GSConv significantly improves the model's ability to capture fine details and subtle features. The enhanced feature extraction capabilities of GSConv enable more precise identification and classification of these small objects, which is vital for effective monitoring and conservation.

The ECA mechanism strengthen the model's ability to enhance feature representation and fusion (Wang et al., 2020). In this study, the ECA mechanism is designed to improve feature fusion and recognition, making it effective for the task of detecting hermit crabs in UAV imagery. ECA consists of a channel attention module and a spatial attention module. The channel attention module utilizes a channel shuffle unit to learn feature dependencies and a SoftMax function to determine channel weights, while the spatial attention module employs a group normalization layer to learn spatial dependencies and a sigmoid function to assign spatial weights. This mechanism enables the network to focus more on the small-scale features that are crucial for recognizing hermit crabs, thereby enhancing detection accuracy and reliability.

### 2.5.3. Evaluation metrics for object detection

Evaluating the performance of object detection deep learning models involves several metrics, including Precision, Recall, F1-score and mAP (Géron, 2019; Everingham et al., 2010). Precision assesses the accuracy of the model's positive predictions by measuring how many of the detected hermit crabs are correctly identified, crucial for minimizing false alarms in ecological monitoring. Recall evaluates the model's ability to detect all actual hermit crabs, ensuring that most or all hermit crabs present in the images are accurately identified. Meanwhile, mAP combines both precision and recall across all classes, providing a comprehensive overview of the model's effectiveness in detecting hermit crabs in various environments, thus supporting accurate and reliable ecological studies. Table 2 shows the confusion matrix between ground truth and prediction.

The evaluation metrics of precision, recall and mAP can be calculated by Equations (4)–(7) respectively:

$$P = \frac{TP}{TP + FP} \tag{4}$$

$$R = \frac{TP}{TP + FN} \tag{5}$$

$$F1 - score = \frac{2 \times P \times R}{P + R} \tag{6}$$

$$mAP = \frac{1}{N} \sum_{i=1}^{N} \left( \sum_{j=0}^{n-1} (r_i - r_j) p_{interp} (r_{j+1}) \right) \tag{7}$$

where $TP_1$ refers to the number of pixels correctly identified as underwater hermit crabs, while $FP_1$ denotes the number of non-hermit crab objects mistakenly identified as underwater hermit crabs. $FP_2$ represents background pixels incorrectly classified as underwater hermit crabs. Similarly, $TP_2$ is the number of pixels accurately detected as hermit crabs on sand, and $FP_3$ is the number of non-hermit crab objects incorrectly classified as hermit crabs on sand. $TN$ indicates the pixels correctly identified as background, $FN_1$ denotes underwater hermit crabs misclassified as non-hermit crab objects, $FN_2$ represents hermit crabs on sand misclassified as non-hermit crab objects, and, $FN_3$ denotes hermit crabs misclassified as background. The F1-score, which is the harmonic mean of precision and recall, provides a balanced measure of a model's performance. In this study, N signifies the total number of distinct object categories identified, which is 2 due to the detection of both underwater hermit crabs and hermit crab on land entities. The parameter n represents the number of recall stages for initial interpolation precision, sequenced in an ascending manner. The terms r and p are used to denote recall and precision, respectively. Compared to the F1-score, the mAP offers a superior evaluation metric for models detecting multiple classes. While the F1-score provides a singular metric assessment, mAP delivers a comprehensive evaluation of the model's efficacy across various categories. This makes mAP a more favorable metric for assessing object detection models that categorize multiple types of objects.

## 2.6. Evaluation and density mapping

This paper employs control variable method to evaluate and select the most suitable SRR network for hermit crab object detection. The methodology includes using SR images reconstructed by various SRR networks as the testing dataset for CRAB-YOLO, with mAP serving as the evaluation index. The study also systematically evaluates various object detection networks using the same SR testing dataset, comparing their mAP results to determine the optimal choice for hermit crabs object detection. Understanding the density distribution of hermit crabs helps identify areas of ecological stress or degradation, indicating the presence of pollutants, habitat loss, or climate change effects. This information supports sustainable fisheries management by enabling targeted conservation measures and protecting critical habitats, ensuring long-term sustainability of marine resources. Additionally, dynamic density distribution data contribute to broader ecological models and climate studies, aiding in global efforts to mitigate climate change impacts on marine ecosystems. Therefore, this study uses the CRAB-YOLO model's detection results to generate a comprehensive distribution map of hermit crabs within the surveyed area.

**Table 2**
The confusion matrix between ground truth and prediction.

| | | Predicted by model | | | |
|---|---|---|---|---|---|
| | | Underwater hermit crabs | Hermit crabs on sand | Background | Sum |
| Ground Truth | Underwater hermit crabs | True Positive ($TP_1$) | False Positive ($FP_1$) | False Positive ($FP_2$) | Predicted Underwater Hermit Crabs ($TP_1 + FP_1 + FP_2$) |
| | Hermit crabs on sand | False Negative($FN_1$) | True Positive ($TP_2$) | False Positive($FP_3$) | Predicted Hermit crabs on sand($FN_1 + TP_2 + FP_3$) |
| | Background | False Negative($FN_2$) | False Negative($FN_3$) | True Negative (TN) | Predicted Background ($FN_2 + FN_3 +$ TN) |
| | Sum | Actual Underwater Crabs($TP_1 + FN_1 + FN_2$) | Actual Hermit Crabs on sand($FP_1 + TP_2 + FN_3$) | Actual Background($FP_2 + FP_3 + TN$) | $TP_1 + TP_2 + TN + FP_1 + FP_2 + FP_3 + FN_1 + FN_2 + FN_3$ |

## 3. Experiment and results

### 3.1. Experiment setup

First, an experiment was designed to evaluate the performance of five SRR models—SRCNN, EDSR, RDN, RCAN, and SRFBN—in enhancing the resolution of LR images. The experimental process involved training all SRR models on the same dataset and then applying them to a new LR test set to reconstruct SR images. The quality of the reconstructed images generated by each SRR network was evaluated using metrics such as Peak PSNR and SSIM. Visual comparisons were also conducted to validate the quantitative evaluation results. In the second part of the experiment, the CRAB-YOLO network developed in this study was trained using HR images. Subsequently, the network was employed to detect hermit crabs using HR, bicubic, and SR images. The detection performance was assessed using precision, recall, and mAP metrics.

### 3.2. Results of super-resolution reconstruction

All deep learning-based SRR models outperformed the traditional Bicubic in terms of PSNR and SSIM. This indicates that SR images reconstructed by deep learning networks exhibit less noise and better preserve edges, textures, and other details of hermit crabs, which is beneficial for subsequent object detection tasks. The results of Table 3 suggest that the algorithm improvements meet both machine and human observer standards. Notably, the SRR method based on the dense network achieved higher PSNR and SSIM scores, with the RDN network delivering the best reconstruction results of 39.5 dB PSNR and 86.54 % SSIM at 4× magnification, surpassing Bicubic interpolation by 2.81 dB PSNR and 2.6 % SSIM, respectively.

Fig. 5 illustrates a comparative analysis of the original high-resolution images and their reconstructions using various methods at different magnifications. It demonstrates that line textures in images reconstructed by the Bicubic method significantly differ from those in the original high-resolution images. In contrast, reconstructions using deep learning methods offer clearer and more precise edge and texture details, closely resembling the original images. All the CNN-based methods produced sharper images than the Bicubic method, confirming consistent trends in visual effects and quantitative evaluation indicators.

**Table 3**
Evaluation metrics of various SRR methods on the LR testset with the magnification factor 4.

| Metrics | Bicubic | SRCNN | SRFBN | EDSR | RCAN | RDN |
|---|---|---|---|---|---|---|
| PSNR (dB) | 36.24 | 36.40 | 36.58 | 36.66 | 36.97 | **37.05** |
| SSIM (%) | 83.94 | 85.13 | 85.45 | 85.59 | 86.44 | **86.54** |

While the SRR models demonstrated excellent performance, it is important to note the potential domain gap between these datasets and real-world images. The simulated LR images used for training may not represent the complexities of real-world low-resolution images, which can include variations in noise, lighting, and texture details. This domain gap may affect the generalization of the models when applied to real-world scenarios. As suggested by , domain adaptation techniques or additional fine-tuning may be required to enhance the performance of SRR models in real-world setups. Future research will focus on assessing the performance of SRR models in real-world field conditions and investigating methods to reduce the domain gap, such as through transfer learning or further fine-tuning using real-world datasets.

### 3.3. Object detection of reconstructed hermit crabs images

To evaluate the impact of various modules on the model's ability to handle multi-scale features and expand its receptive field, a series of ablation experiments were performed. These experiments assessed the object detection efficacy of several network configurations: YOLOv8s as the baseline model without modifications, YOLOv8-1 with an added detection head, YOLOv8-1-2 incorporating the GSconv module and an extra detection head compared to the original YOLOv8s, and YOLOv8-1-2-3 featuring the ECA mechanisms in addition to the previous modules. The YOLOv8-1-2-3 model, ultimately adopted and named CRAB-YOLO for further application, demonstrated the most promising results. The object detection outcomes are presented in Table 4. The YOLOv8-1 model showed a 4.3 % improvement in mAP compared to the YOLOv8s baseline. The YOLOv8-1-2 model further improved mAP by 1.4 % over the YOLOv8-1 model. The YOLOv8-1-2-3 model, now referred to as CRAB-YOLO, achieved the best results, with a 5.9 % increase in mAP compared to the original YOLOv8s model. Notably, compared to the YOLOv8-1-2 model, the CRAB-YOLO model exhibited a 0.7 % decrease in recall but a significant 2.7 % increase in precision. This indicates that while the CRAB-YOLO model might miss some instances but not significantly; the substantial increase in precision means that CRAB-YOLO model has higher confidence in identifying hermit crabs. Models with low precision but high recall are disadvantageous for practical benthic ecosystem monitoring because they may incorrectly classify other objects as the target instances, affecting the population density mapping derived from object detection results (Duan et al., 2020). This misclassification restrains accurate assessment of population dynamics and fisheries guidance. CRAB-YOLO enhances model precision and shows substantial improvement in the overall mAP metric, with only a 0.7 % reduction in recall. In addition, the improved YOLOv8 experiences an increase in inference time of approximately 20 % compared to the original version. Despite this increase, both versions of YOLOv8 maintain a processing efficiency exceeding 100 FPS, which is

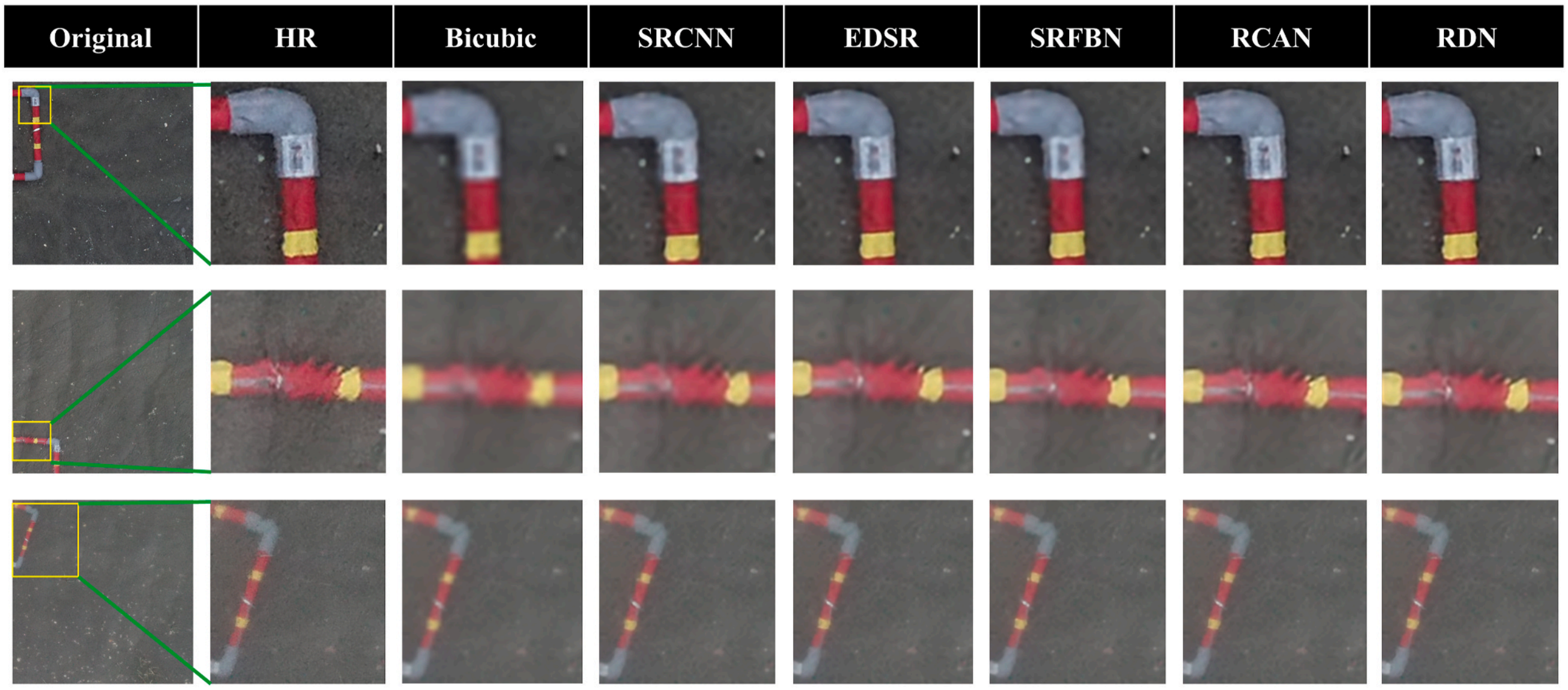

**Fig. 5.** Comparison of the visual effects of reconstructed images based on the six reconstructed methods.

**Table 4**
Ablation experiments of different detection networks.

| model | 4 head | GSConv | ECA | Precision (%) | Recall (%) | mAP@50 (%) |
|---|---|---|---|---|---|---|
| YOLOv8s | × | × | × | 92.3 | 81.1 | 87.2 |
| YOLOv8s-1 | ✓ | × | × | 96.0 | 83.4 | 91.5 |
| YOLOv8s-1-2 | ✓ | ✓ | × | 94.6 | 87.7 | 92.9 |
| YOLOv8s-1-2-3 (CRAB-YOLO) | ✓ | ✓ | ✓ | 97.3 | 87.0 | 93.1 |

*A check mark (✓) indicates the strategy module was used and a cross ( × ) indicates it was not used.

sufficient for handling large datasets and achieving real-time detection. The trade-off between a slight increase in processing time and a substantial gain in accuracy is justified by the practical benefits. The improved YOLOv8 not only maintains high efficiency but also provides more reliable detection results, which is essential for our study. While previous studies, such as Li et al. (2024), reported an mAP of 64.1 % in detecting larger objects like cars using UAV images, our method provides comparable detection accuracy for marine organisms, even in challenging environments with poor water quality. Moreover, Liu et al. (2025) reported an accuracy of 86.7 % for underwater object detection, but their research was conducted in areas with better water quality and fewer environmental constraints. Therefore, our results are significant given the specific conditions of this study. This improvement is meaningful for practical applications as it ensures more reliable detection of hermit crab distributions, thereby enhancing monitoring efforts and the accuracy of density dynamics.

The results depicted in Fig. 6 and Table 5 reveal a strong positive correlation between the PSNR and SSIM of the image reconstruction effect and the hermit crab target detection accuracy, as measured by the average confidence (AC) score. In object detection, the confidence score quantifies the model's certainty about the presence of an object within a detected region. Expressed as a probability value ranging from 0 to 1, a higher score signifies greater confidence in the classification of the detected object (Maji et al., 2022).

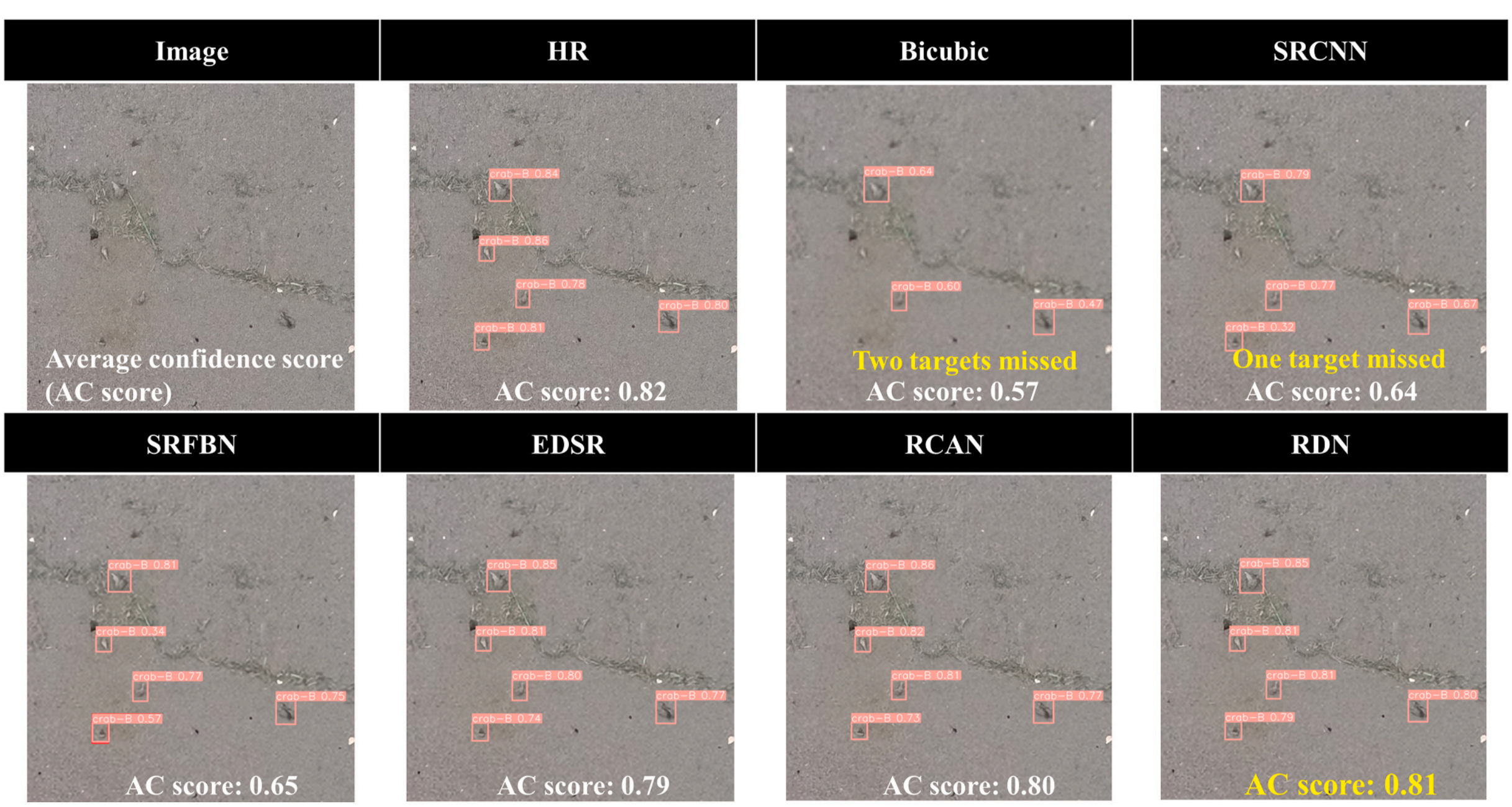

**Fig. 6.** Qualitative comparison of hermit crab detection results of various test sets using CRAB-YOLO.

**Table 5**
Evaluation metrics of detection results of the testset reconstructed by different SRR algorithms.

| Metrics (%) | Bicubic | SRCNN | SRFBN | EDSR | RCAN | RDN |
|---|---|---|---|---|---|---|
| Precision | 38.5 | 69.4 | 72.4 | 70.1 | 68.9 | **73.2** |
| Recall | 20.8 | 47.5 | 56.0 | 57.0 | **64.3** | 60.9 |
| mAP@50 | 29.8 | 56.3 | 61.0 | 62.7 | 69.3 | **69.5** |

Unlike the traditional Bicubic method, deep learning-based methods emphasize the reconstruction of high-frequency details, which lead to sharper and more realistic textures in the reconstructed images. Such detailed reconstructions are crucial for identifying tiny objects like hermit crabs in UAV images. Higher PSNR and SSIM values correlate with better detection performance on the super-resolution images produced by respective SRR networks. This correlation is evident in the superior performance of RCAN and RDN, which achieve the highest AC scores as well as PSNR and SSIM values compared to other reconstruction models.

Table 5 presents experimental results that lead to two significant conclusions. Firstly, the CRAB-YOLO model exhibits high sensitivity to PSNR and SSIM, as its accuracy in detecting targets in images reconstructed by various SRR networks closely aligns with the quantitative metrics assessing each network's reconstruction quality. This correlation demonstrates CRAB-YOLO's effectiveness and adaptability in detecting hermit crabs in drone-captured images. The model is capable of capturing critical details in reconstructed images, significantly enhancing detection accuracy, even with minor quality improvements. Secondly, modest improvements in SRR quality can result in considerable enhancements in target detection performance. For instance, an increase of only about 2 dB in PSNR and 2 % in SSIM for the RDN compared to the Bicubic model led to nearly a 40 % improvement in CRAB-YOLO's detection metrics for hermit crabs. This disparity suggests that traditional SRR evaluation metrics may not fully capture the potential gains in model performance. Instead, practical applications such as object detection offer a more comprehensive assessment of a model's effectiveness.

## 4. Discussion

### *4.1. Discussion on the optimal magnification factor*

The SRR magnification factor is crucial as it determines the extent of resolution enhancement from LR images to SR images. Identifying the optimal magnification factor is essential for effectively applying SRR technology in practical applications, impacting the overall quality and accuracy of detection (Yue et al., 2016). The purpose of testing different magnifications in this study was to compare the effects of SRR enhancement on target detection at various magnification factors. This experiment is to explore the optimal magnification for improving the quality of LR images and assess how SR can be used to effectively enhance detection accuracy in practical scenarios. Images captured by UAV were pre-processed to a resolution of 640 × 640 pixels, forming the HR hermit crab dataset. To create the x1-LR (160× 160) training set, HR images were downsampled by a factor of 4 to simulate LR images commonly encountered in practical UAV-based remote sensing applications. The x1-LR test set was used to represent the low-resolution, fuzzy dataset collected in real-world scenarios, on which the deep learning-based SRR was applied. The enhanced images were then magnified by factors of 2×, 3x, 4x, and 5x to assess the effectiveness of RDN in improving detection accuracy. The HR set used for training the SRR-× 2 model was the same HR dataset mentioned in chapter 2.3. The LR dataset was downsampled by the factor of 2 from the HR dataset according to the degradation model. Both the HR dataset and the LR dataset with a factor of 2 were fed into RDN to train the SRR-× 2 model. The × 1-LR test set (160× 160) was input into to SRR-× 2 model to generate × 2-SR test set (320× 320). Similarly, × 3-SR (480× 480), × 4-SR (640× 640), and × 5-SR (800× 800) test sets were generated using respective magnification factors. The performance of the CRAB-YOLO model was evaluated on these SR datasets, with results presented in Table 6. Fig. 7 illustrates the detection results of some reconstructed hermit crab images with different magnification factors.

Fig. 7 demonstrates the detection results of reconstructed hermit crab images across different magnification factors. In the LR image, only a few hermit crabs are detected. As the magnification increases from × 1 to × 4, the number of detected hermit crabs by CRAB-YOLO progressively increases. The highest clarity and detail are observed at × 4-SR, where the model captures more accurate object boundaries and finer details, leading to an increase in the number of correctly detected hermit crabs. Conversely, at x5-SR, detection quality deteriorates, resulting in a higher number of missed hermit crabs.

The precision improves from 39.8 % at × 1-LR to 73.2 % at × 4-SR while drops to 62.1 % at × 5-SR. Recall shows a significant increase from 19.9 % at × 1-LR to 60.9 % at × 4-SR, but also declines to 42.6 % at × 5-SR. The mAP metric follows a similar trend, rising from 18.1 % at x1-LR to 69.5 % at × 4-SR and then falling to 51.0 % at × 5-SR. Performance metrics initially improve with increasing magnification, reaching a peak at 4 × -SR. At lower magnifications, SR images enhance details, making them closer to HR images. However, beyond 4× magnification, the model struggles to simulate realistic textures, resulting in reduced object detection performance. This is because after × 4, the model needs to infer more detailed information when generating high-resolution images, which does not exist in the original low-resolution images. This excessive inference can easily lead to the generation of unrealistic textures and artifacts, thereby reducing the visual quality and practicality of the image. Moreover, higher magnifications increase computational complexity, making the model more susceptible to noise and errors during processing, further exacerbating image quality degradation (Xiong et al., 2019). Therefore, the × 4 factor can effectively balance detail enhancement and computational efficiency. In the related research on SRR in computer vision, various experiments on magnification have obtained similar conclusions to this study. Specifically, SRR models output with a magnification of around 3.5x-4.0x has better performance in advanced tasks such as target detection and segmentation and can balance computing resources and model accuracy (Xiang et al., 2022; Timofte et al., 2018; Li et al., 2023).

### *4.2. Effect of the adopted object detection network on SRR*

This section evaluates the effectiveness of the CRAB-YOLO model with Faster R-CNN, SSD, and other YOLO series networks in object detection tasks. All models were trained using identical HR dataset and tested across HR, Bicubic and SR testsets, as shown in Table 7. Among them, CRAB-YOLO outperforms all other networks on these three test sets. Furthermore, all networks achieved the highest mAP on the HR test set and the lowest on the Bicubic test set. While Faster R-CNN and SSD performed competently, YOLOv8 and CRAB-YOLO exhibited superior accuracy, with CRAB-YOLO achieving a mAP of 69.5 % on the SR test set, surpassing YOLOv8's 66.2 % and other networks. CRAB-YOLO's ability to provide detailed location data of detected objects proves invaluable for ecological research, enabling comprehensive analyses like population density estimations and understanding species distribution factors. This accuracy is critical for monitoring species behavior and assessing environmental changes, enhancing conservation efforts and sustainable management of coastal ecosystems. Table 6 underscores CRAB-YOLO's excellence in object detection and its significance in ecological data analysis, making it a valuable tool for ecological studies and practical applications in environmental protection. To further investigate whether the SR method enhances detection performance beyond simulated LR imagery, we applied SR processing to real-world aerial images captured at 10m altitude. The results demonstrate a significant improvement in detection performance, with the mAP increasing from 34.8 % (original resolution) to 52.4 % (after SR enhancement). This experiment confirms that SR provides practical benefits not only for simulated LR inputs but also for authentic, field-collected high-resolution imagery. To comprehensively evaluate our approach, we additionally examined the performance of an LR-trained YOLO model using LR images as input. Through five independent training runs, this configuration achieved an average mAP of 37.8 %. In contrast, our super-resolution enhanced method demonstrated a remarkable improvement, with a 31.7 % increase in mAP (69.5 % vs 37.8 %). This substantial enhancement effectively confirms that our solution successfully mitigates the feature degradation inherent in low-quality imagery.

**Table 6**
Evaluation metrics of detection results of the testset reconstructed by different magnification factors.

| Metrics (%) | × 1-LR | × 2-SR | × 3-SR | × 4-SR | × 5-SR |
|---|---|---|---|---|---|
| Precision | 39.8 | 61.7 | 68.0 | **73.2** | 62.1 |
| Recall | 19.9 | 46.9 | 48.0 | **60.9** | 42.6 |
| mAP | 18.1 | 50.5 | 55.3 | **69.5** | 51.0 |

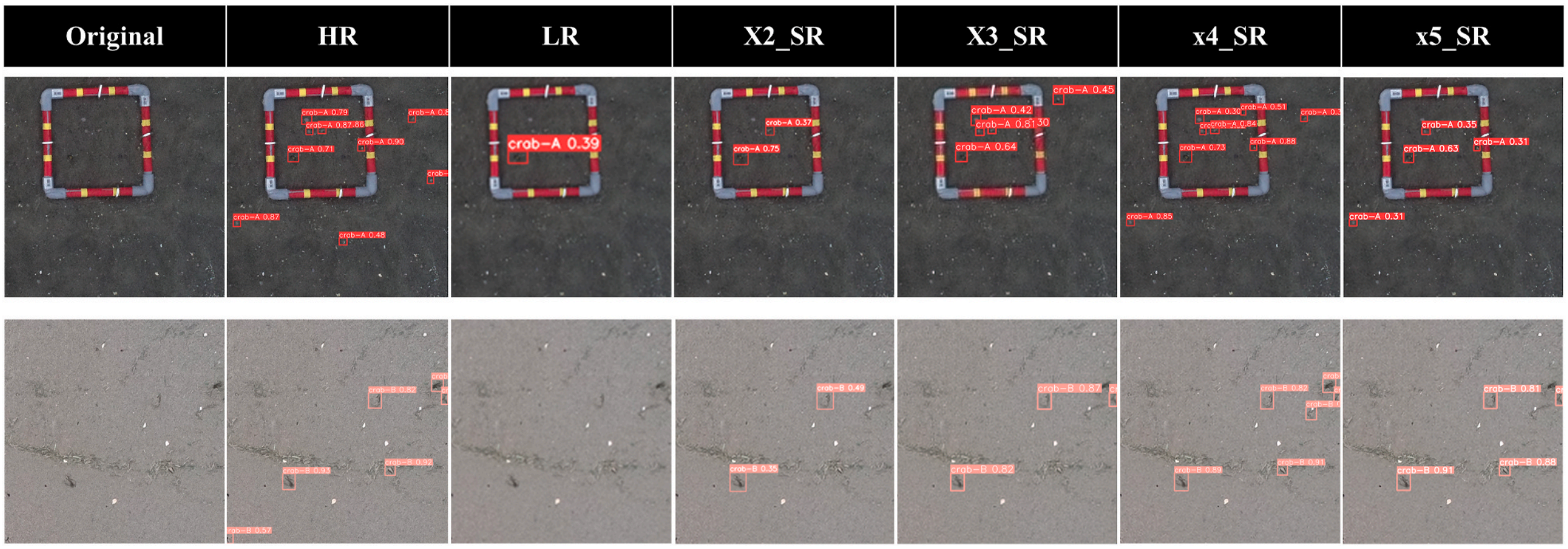

**Fig. 7.** Detection results of hermit crab images from SRR testsets obtained with different magnification factors.

**Table 7**
The detection results of competitive networks on various testsets.

| Different Networks | mAP@50 (%) | | Computational Cost | | |
|---|---|---|---|---|---|
| | Bicubic | HR | SR | Params | Size (M) |
| Faster R-CNN | 12.4 | 80.8 | 52.4 | 28.5M | 113.5 |
| SSD | 11.8 | 71.2 | 47.7 | 26.3M | 95.5 |
| YOLOv3tiny | 6.7 | 86.8 | 49.7 | 8.7M | 17.4 |
| YOLOv4 | 15.1 | 86.2 | 58.1 | 64.4M | 256.3 |
| YOLOv5s | 26.8 | 72.4 | 54.4 | 7.0M | 14.5 |
| YOLOv5Lite-g | 18.9 | 71.5 | 47.7 | 5.5M | 11.3 |
| YOLOX-nano | 19.4 | 69.3 | 49.6 | 0.9M | 3.9 |
| YOLOv7tiny | 16.3 | 72.4 | 52.3 | 6.0M | 12.3 |
| YOLOv8n | 27.7 | 83.2 | 66.2 | 3.0M | 6.2 |
| YOLOv9 | 28.9 | 88.7 | 67.5 | 50.7M | 102.8 |
| **CRAB-YOLO** | **29.8** | **93.1** | **69.5** | 12.9M | 20.5 |

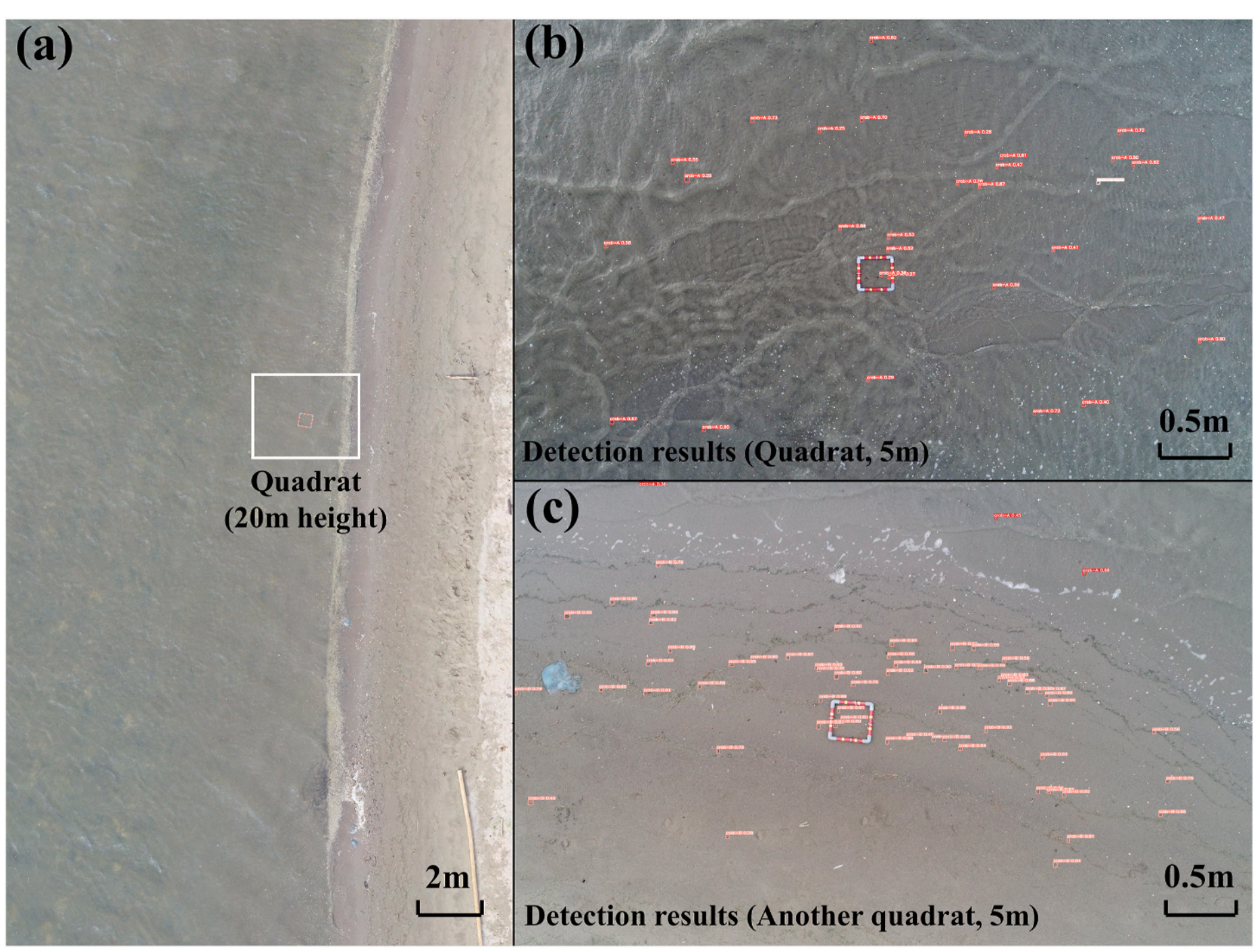

**Fig. 8.** Mapping of hermit crab detection results using the CRAB-YOLO object detection model.

### 4.3. Mapping of detection results based on CRAB-YOLO

Fig. 8 (a) displays an orthophoto taken from a height of 20 m, covering a predefined grid area on the beach. Fig. 8 (b) and (c) illustrate the detection results of the CRAB-YOLO model from a closer distance of 5 m, highlighting the presence of hermit crabs within the detection boxes. The red boxes in these figures mark the successfully detected hermit crabs, demonstrating the model's efficacy in identifying and mapping their distribution. According to the experimental results, images output by the SRR network using RDN were selected and integrated with the CRAB-YOLO model to comprehensively map hermit crab detection in drone images. This high-resolution imagery serves as valuable input for further ecological analyses, allowing for precise identification of hermit crab habitats and generation of detailed maps. Future surveys can utilize UAVs to map benthos habitats from higher altitudes, potentially replacing traditional low-altitude monitoring. While lower flights yield high-quality, high-resolution images, they increase survey time and pose challenges to UAV endurance. The introduction of SRR technology significantly enhances the resolution of high-altitude UAV images, improving the accuracy of object detection within these images. Fig. 5 demonstrates how the RDN algorithm effectively reconstructs four-times downsampled LR images—assumed to be blurry images typically captured from realistic high altitudes scenarios —into clearer SR images. The accuracy of object detection in these SR.

images closely approach that of the ground truth HR images. Adopting SRR for high-altitude surveys could greatly increase efficiency and reduce the costs of ecological studies. Additionally, combining ordinary cameras on UAVs with SRR technology mitigates the need for high-precision cameras, illustrating that SRR algorithms can overcome hardware limitations, thus lowering both the equipment costs and barriers to ecological monitoring. This further validates the practicality and significance of SRR technology in this field. As indicated in Fig. 8, the spatial distribution and density of hermit crabs are clearly visualized, closely aligning with the coastline. The CRAB-YOLO results not only enhance the accuracy of hermit crab detection but also provide crucial data for ecological monitoring. Utilizing computer vision technology, this method overcomes the constraints of traditional survey techniques, offering a robust and scalable solution for real-time assessment and protection of marine ecosystems. Its large-scale visualization capability provides a clear depiction of the ecological status of hermit crabs, supporting decision-making for various stakeholders.

## 5. Conclusion

Hermit crabs are pivotal to coastal ecosystems, playing roles in seed dispersal, organic matter decomposition, and soil disturbance. They are also vital indicators of environmental health due to their sensitivity to pollutants. To address the challenges in hermit crab surveys, such as motion blur and inadequate resolution during UAV-based image acquisition, this study introduces a large-scale survey approach using UAVs and deep learning-based SRR with the self-proposed CRAB-YOLO network. Deep learning-based SRR techniques notably enhance UAV image quality beyond the traditional bicubic method, both in quantitative metrics and visual fidelity, enabling more precise detection and quantification of hermit crab features. Among the experimental groups, the RDN method stands out for its exceptional image enhancement capabilities.

The CRAB-YOLO model demonstrates exceptional accuracy and mAP across different SR datasets, significantly outperforming other detection models. Specifically, the RDN method, when combined with CRAB-YOLO, enhances hermit crab detection by nearly 40 % compared to the bicubic method, making it a highly effective tool for ecological monitoring and conservation efforts. Object detection maps generated from this model clearly illustrate the density distribution of hermit crabs in the surveyed area, highlighting the practicality and applicability of the technology. Given the versatility of deep learning-based SRR and YOLO networks, this survey method could be adapted for other marine benthic organisms like sea cucumbers, corals, and marine debris, potentially paving the way for comprehensive multi-target marine benthic population surveys.

However, certain limitations exist, such as UAV performance being impacted by water quality and the nocturnal nature of some marine organisms, which complicates density assessments. Future enhancements could involve integrating underwater drones for improved image quality or combining UAVs with sonar for all-weather capabilities. Additionally, future iterations of the YOLO series models are expected to further enhance the accuracy of the target detection network. Furthermore, the benthos detection method proposed requires two separate stages, which may not meet the needs for real-time monitoring. Future research will explore an optimized network that integrates super-resolution reconstruction and object detection into a single process.

## CRediT authorship contribution statement

**Fan Zhao:** Writing – review & editing, Writing – original draft, Visualization, Validation, Software, Resources, Project administration, Methodology, Investigation, Funding acquisition, Formal analysis, Data curation, Conceptualization. **Yijia Chen:** Writing – review & editing, Data curation, Conceptualization. **Dianhan Xi:** Validation. **Yongying Liu:** Writing – original draft, Validation. **Jiaqi Wang:** Writing – review & editing, Visualization, Software. **Shigeru Tabeta:** Writing – review & editing, Supervision, Resources. **Katsunori Mizuno:** Writing – review & editing, Supervision, Resources, Conceptualization.

## Declaration of competing interest

The authors declare that they have no known competing financial interests or personal relationships that could have appeared to influence the work reported in this paper.

## Acknowledgements

We extend our heartfelt gratitude to the students from the Department of Environment Systems at The University of Tokyo, the staff of Windy Network Corp., and the Hamanako Branch of the Shizuoka Prefectural Research Institute of Fishery for their invaluable cooperation in the field surveys. This research was partially supported by the Japan Science and Technology Agency SPRING Program (Grant Number JPMJSP2108) and the Grant-in-Aid for Scientific Research (KAKENHI) (Grant Number 20KK0238) from the Japan Society for the Promotion of Science (JSPS). Additional funding was provided by JST PRESTO, Grant Number JPMJPR24G9, Japan.

## Data availability

Data will be made available on request.

## References

Asakura, A., 2021. Crustaceans in changing climate: global warming and invasion of tropical land hermit crabs (Crustacea: *decapoda*: anomura: *coenobitidae*) into temperate area in Japan. Zoology 145, 125893. https://doi.org/10.1016/j.zool.2021.125893.

Bae, H., Jang, K., An, Y.K., 2021. Deep super resolution crack network (SrcNet) for improving computer vision–based automated crack detectability in in situ bridges. Struct. Health Monit. 20 (4), 1428–1442. https://doi.org/10.1177/14759217209172.

Benecki, P., Kawulok, M., Kostrzewa, D., Skonieczny, L., 2018. Evaluating super-resolution reconstruction of satellite images. Acta Astronaut. 153, 15–25. https://doi.org/10.1016/j.actaastro.2018.07.035.

Brehmer, P., Vercelli, C., Gerlotto, F., Sanguinède, F., Pichot, Y., Guennégan, Y., et al., 2006. Multibeam sonar detection of suspended mussel culture grounds in the open sea: direct observation methods for management purposes. Aquaculture 252 (2–4), 234–241. https://doi.org/10.1016/j.aquaculture.2005.06.035.
Briffa, M., Arnott, G., Hardege, J.D., 2023. Hermit crabs as model species for investigating the behavioural responses to pollution. Sci. Total Environ., 167360. https://doi.org/10.1016/j.scitotenv.2023.167360.
Brown, C.J., Beaudoin, J., Brissette, M., Gazzola, V., 2019. Multispectral multibeam echo sounder backscatter as a tool for improved seafloor characterization. Geosciences 9 (3), 126. https://doi.org/10.3390/geosciences9030126.
Chia, Kai Yuan, Chin, Cheng Siong, See, Simon, 2023. Deep transfer learning application for intelligent marine debris detection. In: International Conference on Engineering Applications of Neural Networks. Springer Nature Switzerland, Cham. https://doi.org/10.1007/978-3-031-34204-2_39.
Colefax, A.P., Butcher, P.A., Kelaher, B.P., 2018. The potential for unmanned aerial vehicles (UAVs) to conduct marine fauna surveys in place of manned aircraft. ICES (Int. Counc. Explor. Sea) J. Mar. Sci. 75 (1), 1–8. https://doi.org/10.1093/icesjms/fsx100.
de la Haye, K.L., Spicer, J.I., Widdicombe, S., Briffa, M., 2012. Reduced pH sea water disrupts chemo-responsive behaviour in an intertidal crustacean. J. Exp. Mar. Biol. Ecol. 412, 134–140. https://doi.org/10.1016/j.jembe.2011.11.013.
Dhillon, A., Verma, G.K., 2020. Convolutional neural network: a review of models, methodologies and applications to object detection. Progress in Artificial Intelligence 9 (2), 85–112. https://doi.org/10.1007/s13748-019-00203-0.
Duan, H., Wang, S., Guan, Y., 2020. Sofa-net: second-Order and first-order attention network for crowd counting. arXiv preprint arXiv:2008.03723. https://doi.org/10.48550/arXiv.2008.03723.
Everingham, M., Van Gool, L., Williams, C.K., Winn, J., Zisserman, A., 2010. The pascal visual object classes (voc) challenge. Int. J. Comput. Vis. 88, 303–338. https://doi.org/10.1007/s11263-009-0275-4.
Fu, X., Liu, Y., Liu, Y., 2022. A case study of utilizing YOLOT based quantitative detection algorithm for marine benthos. Ecol. Inform. 70, 101603. https://doi.org/10.1016/j.ecoinf.2022.101603.
González, D., Patricio, M.A., Berlanga, A., Molina, J.M., 2022. A super-resolution enhancement of UAV images based on a convolutional neural network for mobile devices. Personal Ubiquitous Comput. 26 (4), 1193–1204. https://doi.org/10.1007/s00779-019-01355-5.
Geraeds, M., van Emmerik, T., de Vries, R., bin Ab Razak, M.S., 2019. Riverine plastic litter monitoring using unmanned aerial vehicles (UAVs). Remote Sens. 11 (17), 2045. https://doi.org/10.3390/rs11172045.
Géron, A., 2019. Hands-on Machine Learning with Scikit-Learn, Keras, and Tensorflow: Concepts, Tools, and Techniques to Build Intelligent Systems. O'Reilly Media: Sebastopol, CA, USA, 2019. https://www.clc.hcmus.edu.vn/wpcontent/uploads/2017/11/Hands_On_Machine_Learning_with_Scikit_Learn_and_TensorFlow.pdf.
Jiang, P., Ergu, D., Liu, F., Cai, Y., Ma, B., 2022. A review of Yolo algorithm developments. Procedia Comput. Sci. 199, 1066–1073. https://doi.org/10.1016/j.procs.2022.01.135.
Jin, Y., Zhang, Y., Cen, Y., Li, Y., Mladenovic, V., Voronin, V., 2021. Pedestrian detection with super-resolution reconstruction for low-quality image. Pattern Recogn. 115, 107846. https://doi.org/10.1016/j.patcog.2021.107846.
Kieu, H.T., Pak, H.Y., Trinh, H.L., Pang, D.S.C., Khoo, E., Law, A.W.K., 2023. UAV-based remote sensing of turbidity in coastal environment for regulatory monitoring and assessment. Mar. Pollut. Bull. 196, 115482. https://doi.org/10.1016/j.marpolbul.2023.115482.
Lavers, J.L., Sharp, P.B., Stuckenbrock, S., Bond, A.L., 2020. Entrapment in plastic debris endangers hermit crabs. J. Hazard Mater. 387, 121703. https://doi.org/10.1016/j.jhazmat.2019.121703.
Li, C., Li, L., Jiang, H., Weng, K., Geng, Y., Li, L., Wei, X., 2022. YOLOv6: a single-stage object detection framework for industrial applications. arXiv preprint arXiv: 2209.02976. https://doi.org/10.48550/arXiv.2209.02976.
Li, Y., Li, Q., Pan, J., Zhou, Y., Zhu, H., Wei, H., Liu, C., 2024. Sod-yolo: Small-object-detection algorithm based on improved yolov8 for uav images. Remote Sens. 16 (16), 3057.
Li, Y., Zhang, Y., Timofte, R., Van Gool, L., Yu, L., Li, Y., et al., 2023. NTIRE 2023 challenge on efficient super-resolution: methods and results. In: Proceedings of the IEEE Conference on Computer Vision and Pattern Recognition (CVPR), pp. 1921–1959. https://doi.org/10.1109/CVPRW59228.2023.00189.
Li, Z., Yang, J., Liu, Z., Yang, X., Jeon, G., Wu, W., 2019. Feedback network for image super-resolution. In: Proceedings of the IEEE/CVF Conference on Computer Vision and Pattern Recognition (CVPR), pp. 3867–3876. https://doi.org/10.1109/CVPR.2019.00399.
Liu, J., Zhou, R., Li, Y., Ren, P., 2025. Enhanced underwater object detection with YOLO-LDFE: a model for improved accuracy with balanced efficiency. Journal of Real-Time Image Processing 22 (2), 58.
Liu, Y., Wang, S., 2021. A quantitative detection algorithm based on improved faster R-CNN for marine benthos. Ecol. Inform. 61, 101228. https://doi.org/10.1016/j.ecoinf.2021.101228.
Liu, Y., Yeoh, J.K., Chua, D.K., 2020. Deep learning–based enhancement of motion blurred UAV concrete crack images. J. Comput. Civ. Eng. 34 (5), 04020028. https://doi.org/10.1061/(ASCE)CP.1943-5487.0000907.
Mahapatra, D., Bozorgtabar, B., Garnavi, R., 2019. Image super-resolution using progressive generative adversarial networks for medical image analysis. Comput. Med. Imag. Graph. 71, 30–39. https://doi.org/10.1016/j.compmedimag.2018.10.005.
Maji, D., Nagori, S., Mathew, M., Poddar, D., 2022. Yolo-pose: enhancing YOLO for multi-person pose estimation using object keypoint similarity loss. In: Proceedings of the IEEE Conference on Computer Vision and Pattern Recognition. CVPR), pp. 2637–2646. https://doi.org/10.1109/cvprw56347.2022.00297.
Matarrese, A., Mastrototaro, F., D'onghia, G., Maiorano, P., Tursi, A., 2004. Mapping of the benthic communities in the Taranto seas using side-scan sonar and an underwater video camera. Chem. Ecol. 20 (5), 377–386. https://doi.org/10.1080/02757540410001727981.
Nafchi, M.A., Chamani, A., 2019. Physiochemical factors and heavy metal pollution, affecting the population abundance of Coenobita scaevola. Mar. Pollut. Bull. 149, 110494. https://doi.org/10.1016/j.marpolbul.2019.110494.
Ooi, Y.K., Ibrahim, H., 2021. Deep learning algorithms for single image super-resolution: a systematic review. Electronics 10 (7), 867. https://doi.org/10.3390/electronics10070867.
Rasti, P., Uiboupin, T., Escalera, S., Anbarjafari, G., 2016. Convolutional neural network super resolution for face recognition in surveillance monitoring. In: Articulated Motion and Deformable Objects: 9Th International Conference, AMDO 2016, Palma De Mallorca, Spain, July 13-15, 2016, Proceedings, pp. 175–184. https://doi.org/10.1007/978-3-319-41778-3_18, 9.
Redmon, J., Divvala, S., Girshick, R., Farhadi, A., 2016. You only look once: unified, real-time object detection. Proceedings of the IEEE conference on computer vision and pattern recognition (CVPR) 2016, 779–788. https://doi.org/10.1109/CVPR.2016.91.
Sant'Anna, B.S., Dos Santos, D.M., Sandron, D.C., De Souza, S.C., De Marchi, M.R.R., Zara, F.J., Turra, A., 2012. Hermit crabs as bioindicators of recent tributyltin (TBT) contamination. Ecol. Indic. 14 (1), 184–188. https://doi.org/10.1016/j.ecolind.2011.08.010.
Shen, L., Lang, B., Song, Z., 2023. DS-YOLOv8-based object detection method for remote sensing images. IEEE Access 11, 125122–125137. https://doi.org/10.1109/access.2023.3330844.
Shih, C.C., Horng, M.F., Tseng, Y.R., Su, C.F., Chen, C.Y., 2019. An adaptive bottom tracking algorithm for side-scan sonar seabed mapping. In: 2019 IEEE Underwater Technology (UT). IEEE, pp. 1–7. https://doi.org/10.1109/ut.2019.8734291.
Song, G., Liu, Y., Wang, X., 2020. Revisiting the sibling head in object detector. In: Proceedings of the IEEE/CVF Conference on Computer Vision and Pattern Recognition (CVPR), pp. 11563–11572. https://doi.org/10.1109/CVPR42600.2020.01158.
Timofte, R., Gu, S., Wu, J., Van Gool, L., 2018. Ntire 2018 challenge on single image super-resolution: methods and results. In: Proceedings of the IEEE Conference on Computer Vision and Pattern Recognition (CVPR), pp. 852–863. https://doi.org/10.1109/CVPRW.2018.00130.
Turra, A., Ragagnin, M.N., McCarthy, I.D., Fernandez, W.S., 2020. The effect of ocean acidification on the intertidal hermit crab *Pagurus criniticornis* is not modulated by cheliped amputation and sex. Mar. Environ. Res. 153, 104794. https://doi.org/10.1016/j.marenvres.2019.104794.
Ventura, D., Bonifazi, A., Gravina, M.F., Belluscio, A., Ardizzone, G., 2018. Mapping and classification of ecologically sensitive marine habitats using unmanned aerial vehicle (UAV) imagery and object-based image analysis (OBIA). Remote Sens. 10 (9), 1331. https://doi.org/10.3390/rs10091331.
Wang, J., Zhao, F., Shao, X., Liu, Y., Xi, D., Ma, B., Mizuno, K., 2024. Super-resolution approaches based shallow-water benthic identification using multispectral satellite imagery. In: OCEANS 2024- Halifax, NS, Canada. IEEE, pp. 1–4. https://doi.org/10.1109/OCEANS55160.2024.10754245.
Wang, Q., Wu, B., Zhu, P., Li, P., Zuo, W., Hu, Q., 2020. ECA-Net: efficient channel attention for deep convolutional neural networks. In: Proceedings of the IEEE/CVF Conference on Computer Vision and Pattern Recognition (CVPR), pp. 11534–11542. https://doi.org/10.1109/CVPR42600.2020.01155.
Wang, Z., Liu, D., Yang, J., Han, W., Huang, T., 2015. Deep networks for image super-resolution with sparse prior. In: Proceedings of the IEEE Conference on Computer Vision and Pattern Recognition (CVPR), pp. 370–378. https://doi.org/10.1109/ICCV.2015.50.
Ward, C.M., Harguess, J., Crabb, B., Parameswaran, S., 2017. Image quality assessment for determining efficacy and limitations of super-resolution convolutional neural network (SRCNN). Applications of Digital Image Processing XL. International Society for Optics and Photonics. https://doi.org/10.1117/12.2275157.
Xiang, C., Wang, W., Deng, L., Shi, P., Kong, X., 2022. Crack detection algorithm for concrete structures based on super-resolution reconstruction and segmentation network. Autom. ConStruct. 140, 104346. https://doi.org/10.1016/j.autcon.2022.104346.
Xiong, D., Huang, K., Chen, S., Li, B., Jiang, H., Xu, W., 2019. Noucsr: efficient super-resolution network without upsampling convolution. In: 2019 IEEE/CVF International Conference on Computer Vision Workshop (ICCVW). IEEE, pp. 3378–3387. https://doi.org/10.1109/ICCVW.2019.00420.
Xu, X., Liu, Y., Lyu, L., Yan, P., Zhang, J., 2023. MAD-YOLO: a quantitative detection algorithm for dense small-scale marine benthos. Ecol. Inform. 75, 102022. https://doi.org/10.1016/j.ecoinf.2023.102022.
Yang, H., Wang, Z., Liu, X., Li, C., Xin, J., Wang, Z., 2023. Deep learning in medical image super resolution: a review. Appl. Intell. 53 (18), 20891–20916. https://doi.org/10.1007/s10489-023-04566-9.
Yue, L., Shen, H., Li, J., Yuan, Q., Zhang, H., Zhang, L., 2016. Image super-resolution: the techniques, applications, and future. Signal Process. 128, 389–408. https://doi.org/10.1016/j.sigpro.2016.05.002.
Zhang, Y., Li, K., Li, K., Wang, L., Zhong, B., Fu, Y., 2018a. Image super-resolution using very deep residual channel attention networks. In: Proceedings of the European Conference on Computer Vision. ECCV), pp. 286–301. https://doi.org/10.1007/978-3-030-01234-2_18.
Zhang, Y., Tian, Y., Kong, Y., Zhong, B., Fu, Y., 2018b. Residual dense network for image super-resolution. In: Proceedings of the IEEE Conference on Computer Vision and Pattern Recognition (CVPR), pp. 2472–2481. https://doi.org/10.1109/CVPR.2018.00262.

Zhao, F., Mizuno, K., Tabeta, S., Hayami, H., Fujimoto, Y., Shimada, T., 2022. New method of mussel survey by using high-resolution acoustic video camera-ARIS and deep learning. OCEANS 2022 - Chennai 1–4. https://doi.org/10.1109/OCEANSChennai45887.2022.9775335. Chennai, India, 2022.

Zhao, F., Mizuno, K., Tabeta, S., Hayami, H., Fujimoto, Y., Shimada, T., 2023. Survey of freshwater mussels using high-resolution acoustic imaging sonar and deep learning-based object detection in Lake Izunuma, Japan. Aquat. Conserv. Mar. Freshw. Ecosyst. 1–14. https://doi.org/10.1002/aqc.4040.

Zhao, F., Liu, Y., Wang, J., Chen, Y., Xi, D., Shao, X., Tabeta, S., Mizuno, K., 2024a. Riverbed litter monitoring using consumer-grade aerial-aquatic speedy scanner (AASS) and deep learning based super-resolution reconstruction and detection network. Mar. Pollut. Bull. 209 (Part A), 117030. https://doi.org/10.1016/j.marpolbul.2024.117030, 2024.

Zhao, F., Ren, Z., Wang, J., Wu, Q., Xi, D., Shao, X., Liu, Y., Chen, Y., Mizuno, K., 2024b. Smart UAV-assisted rose growth monitoring with improved YOLOv10 and Mamba restoration techniques. Smart Agricultural Technology 10, 100730. https://doi.org/10.1016/j.atech.2024.100730, 2024.

Zhao, F., Xi, D., Chen, Y., Ma, B., Liu, Y., Wang, J., Mizuno, K., 2024c. Basic study of deep learning based efficient hermit crabs detection from drone-captured images. OCEANS 2024 - Singapore, pp. 1–5. https://doi.org/10.1109/OCEANS51537.2024.10706264. Singapore, Singapore, 2024.

Zhao, H., Zhang, H., Zhao, Y., 2023. Yolov7-sea: object detection of maritime UAV images based on improved YOLOv7. In: Proceedings of the IEEE/CVF Winter Conference on Applications of Computer Vision, pp. 233–238. https://doi.org/10.1109/wacvw58289.2023.00029.

Zhao, X., Zhang, X., 2018. Residual super-resolution single shot network for low-resolution object detection. IEEE Access 6, 47780–47793. https://doi.org/10.1109/access.2018.2867586.

Zhao, F., He, Y., Song, J., et al., 2025a. Smart UAV-assisted blueberry maturity monitoring with mamba-based computer vision. Precis. Agric. 26 (56). https://doi.org/10.1007/s11119-025-10252-2, 2025.

Zhao, F., Huang, B., Wang, J., et al., 2025b. Seafloor debris detection via underwater photogrammetry and deep learning-driven image restoration: a case study from Koh Tao, Thailand. Mar. Pollut. Bull. 214, 117710, 2025.